\documentclass[letterpaper]{article}
\usepackage{proceed2e}
\usepackage[margin=1in]{geometry}
\usepackage{amsmath}
\usepackage{amsfonts}
\usepackage{dsfont}
\usepackage{bm}
\usepackage{bbm}
\newcommand{\argmin}{\operatornamewithlimits{argmin}}
\newcommand{\argmax}{\operatornamewithlimits{argmax}}

\usepackage{graphicx, subfigure} 
\usepackage{caption}

\usepackage{times}

\title{Learning Approximately Objective Priors}

\author{{\bf Eric Nalisnick}  \\
Department of Computer Science        \\
University of California, Irvine  \\
\texttt{enalisni@uci.edu}\\
\And
{\bf Padhraic Smyth} \\
Department of Computer Science\\
University of California, Irvine\\
\texttt{smyth@ics.uci.edu}} 

\begin{document}

\maketitle

\begin{abstract}
Informative Bayesian priors are often difficult to elicit, and when this is the case, modelers usually turn to noninformative or objective priors.  However, objective priors such as the Jeffreys and reference priors are not tractable to derive for many models of interest.  We address this issue by proposing techniques for learning reference prior approximations: we select a parametric family and optimize a black-box lower bound on the reference prior objective to find the member of the family that serves as a good approximation.  We experimentally demonstrate the method's effectiveness by recovering Jeffreys priors and learning the Variational Autoencoder's reference prior. 
\end{abstract}

\section{INTRODUCTION}
Bayesian inference is distinguished by its ability to incorporate existing knowledge in a principled way.  This knowledge is encoded by the prior distribution, and its specification is widely regarded as ``the most important step" in the Bayesian approach given that it can ``drastically alter the subsequent inference" \cite{robert2007bayesian}.  The choice is especially crucial in settings with few observations, as the data cannot overwhelm a harmful prior. Even in situations where there are enough observations to make the prior's influence on the posterior benign,
the marginal likelihood is still sensitive to the choice, possibly resulting in the selection of an inferior model. 

The best case scenario for specifying a prior is when---unsurprisingly---there is existing information about the phenomenon we wish to model.  For example, choosing a good prior for parameters in a model of galaxy formation can usually be done by consulting an astrophysicist or the relevant research literature.   

In many practical situations, however, there are no available means for obtaining useful prior information.  For example, in high-dimensional problems the parameter space is often
inherently unintuitive.  The usual way to proceed is to pick a noninformative prior that is \textit{flat} and/or \textit{objective}.  By \textit{flat prior} we mean a distribution that does not have any substantial concentration of its mass; maximum entropy priors \cite{jaynes1957information} often exhibit this characteristic.  An \textit{objective} prior is one that has some formal invariance property.  The two best known examples of objective priors are \textit{Jeffreys} \cite{jeffreys1946invariant} and \textit{reference} \cite{bernardo1979reference} priors, which are both invariant to model reparametrization.  Some priors are both objective and flat: the Jeffreys prior for the Gaussian mean is the (improper) uniform distribution. However, just because a prior is relatively flat does not mean it is objective.  For example, the Bernoulli's Jeffreys prior is the arcsine distribution, which, having vertical asymptotes at $0$ and $1$, is conspicuously not flat.

Since there are no guarantees that what looks to be a flat prior might not harbor hidden subjectivity, objective priors seem to be the better `default' choices.  However, the mathematical rigor that makes objective priors attractive also makes their use problematic: their derivation is difficult for all but the simplest models.  To be specific, solving the calculus of variations problem for a reference prior requires, among other properties, an analytical form for the posterior distribution, which is rarely available.      

In this paper  we broaden the potential use of objective priors by describing methods for learning high-fidelity reference prior approximations.  The proposed method is akin to black-box (posterior) variational inference \cite{ranganath2014black}: we posit a parametric family of distributions and perform derivation-free optimization to find the member of the family closest to the true reference prior.  Doing so would be useful, for example, if one wishes to have an objective prior that preserves model conjugacy\footnote{Reference priors are often improper distributions.}.  The modeler could employ the techniques proposed below to find the conjugate prior's parameter setting that makes it closest to objective.  Moreover, these methods learn a reference prior for a given model {independently} of any data source\footnote{Except when the model is for a conditional distribution, i.e. $p(y | x)$.  In this case, samples of $x$ are necessary to learn the approximate reference prior.}, which means that obtaining a reference prior for a particular model needs to be done only once.  

In our experimental results we demonstrate that the proposed framework recovers the Jeffreys prior better than existing numerical methods.  We also analyze the optimization objective, providing intuition behind a number of hyper-parameter choices.  And lastly, we learn a reference prior for a \textit{Variational Autoencoder} \cite{kingma2013auto}.  In an interesting case study, we see that the Variational Autoencoder's reference prior differs markedly from the standard Normal distribution that is commonly used as the prior on the latent space.  

\section{BACKGROUND AND \\RELATED WORK}
We begin by defining reference priors, highlighting their connection to the Jeffreys prior, and summarizing the related work on computing intractable reference priors.  We use the following notation throughout the paper.  Define the likelihood to be $p(\mathcal{D}|\boldsymbol{\theta}) = \prod_{i=1}^{N}p(\mathbf{x}_{i}|\boldsymbol{\theta})$ where $\boldsymbol{\theta}$ are the model parameters and $\mathcal{D}$ is the dataset, which is comprised of $N$ i.i.d. observations $\mathbf{x}_{i} \in \mathcal{X}$.  $p(\boldsymbol{\theta})$ denotes the prior, $p(\boldsymbol{\theta} | \mathcal{D})$ the posterior, and $p(\mathcal{D})=\int_{\boldsymbol{\theta}} p(\mathcal{D}|\boldsymbol{\theta}) p(\boldsymbol{\theta})d \boldsymbol{\theta}$ the marginal likelihood (or model evidence).  When we refer to the `likelihood function' we mean the functional form of the data model, $p(\mathbf{x}|\boldsymbol{\theta})$.  We write expectations with respect to the dataset likelihood, but because of $\mathcal{D}$'s i.i.d. assumption, these can be written equivalently in terms of each data instance; for example: $\mathbb{H}_{\mathcal{D}|\boldsymbol{\theta}}[\mathcal{D}] = {-}\int_{\mathcal{D}} p(\mathcal{D}|\boldsymbol{\theta}) \log p(\mathcal{D}|\boldsymbol{\theta}) d \mathcal{D} = {-}\int_{\mathbf{x}} \prod_{i}p(\mathbf{x}_{i}|\boldsymbol{\theta}) \log \prod_{i}p(\mathbf{x}_{i}|\boldsymbol{\theta}) d \mathbf{x} =  N \mathbb{H}_{\mathbf{x}|\boldsymbol{\theta}}[\mathbf{x}]$. 

\subsection{REFERENCE PRIORS}
\textit{Reference priors}  \cite{berger2009formal,bernardo2005reference} (RPs) are objective Bayesian prior distributions derived for a given likelihood function by finding the prior that  {maximizes} the data's influence on the posterior distribution.  Equivalently, the prior's influence on the posterior is minimized, which is precisely the behavior we desire if we wish to represent a state of ignorance about the model parameters.  The RP's data-driven nature yields `frequentist-esque' posteriors:
for large sample sizes, the $1-\alpha$ credible interval approximates a confidence interval with significance level $\alpha$ \cite{irony1997non}.  Thus, RPs give results that are the nearest Bayesian equivalent to maximum likelihood estimation, and this behavior is how they derive their name: RPs serve as a \textit{reference} against which to test subjective priors. 

\textbf{Definition.}  We now state the RP definition formally.  A RP $p^{*}(\boldsymbol{\theta})$ is the distribution that maximizes the mutual information between the parameters $\boldsymbol{\theta}$ and the data $\mathcal{D}$ \cite{berger2009formal, bernardo1979reference}: \begin{equation}\label{refPrior}\begin{split}
    p^{*}(\boldsymbol{\theta}) &= \argmax_{p(\boldsymbol{\theta})} \ \  I(\boldsymbol{\theta}, \mathcal{D}) \\ &=  \argmax_{p(\boldsymbol{\theta})}  \underbrace{\mathbb{H}[\boldsymbol{\theta}]}_{\substack{\text{maximize prior} \\ \text{uncertainty}}} - \underbrace{\mathbb{H}[\boldsymbol{\theta} | \mathcal{D}]}_{\substack{\text{minimize posterior} \\ \text{uncertainty}}}. 
\end{split}
\end{equation}  Here $I(\cdot, \cdot)$ denotes mutual information.  In the second line,  $I(\cdot, \cdot)$ is (by definition) decomposed into separate marginal and conditional entropy terms, showing that maximizing $I(\boldsymbol{\theta}, \mathcal{D})$ in turn maximizes the prior's uncertainty while minimizing the posterior's uncertainty.  The second term reflects the RP's data-driven nature as it encourages the posterior to contract quickly (as $N$ increases). 
Another way to see how the RP accentuates the data's influence is by writing the mutual information in terms of a Kullback-Leibler divergence (KLD): $I(\boldsymbol{\theta}, \mathcal{D}) =  \int_{\mathcal{D}}p(\mathcal{D}) \  \text{KLD}[p(\boldsymbol{\theta} | \mathcal{D}) \mid \mid p(\boldsymbol{\theta})] \  d\mathcal{D}$.  This form shows that increasing $I(\boldsymbol{\theta}, \mathcal{D})$ decreases the similarity between the posterior and prior.

\textbf{Solution.}  Solving Equation \ref{refPrior} for $p^{*}$ is a calculus of variations problem whose solution can be expressed by re-writing the mutual information as \begin{equation}\label{refPrior3}\begin{split}
     I(\boldsymbol{\theta}, \mathcal{D}) &=   {- \int_{\boldsymbol{\theta}} p(\boldsymbol{\theta}) \ \log \frac{p(\boldsymbol{\theta})}{f(\boldsymbol{\theta})} \  d\boldsymbol{\theta} }\\ \text{ where } \ f(\boldsymbol{\theta}) = \exp &  \{ \int_{\mathcal{D}} p(\mathcal{D}|\boldsymbol{\theta}) \ \log p(\boldsymbol{\theta} | \mathcal{D}) \ d \mathcal{D} \}.
\end{split}
\end{equation} 
Clearly, the mutual information is maximized when $p(\boldsymbol{\theta}) \propto f(\boldsymbol{\theta})$.  See Bernardo (1979) for a complete discussion of the derivation.  Equation \ref{refPrior3} also makes clear the analytical obstacles that need to be overcome to solve the optimization problem for a given model: the functional $f$ requires that the log posterior---which is usually intractable to compute---be integrated over the likelihood function.  Note that the solution is commonly not a proper distribution (that integrates to 1).

\textbf{Relation to Jeffreys Priors.}  RPs are equal to the Jeffreys\footnote{The Jeffreys prior is defined as $\pi(\boldsymbol{\theta}) \propto \sqrt{\det \mathcal{F}[\boldsymbol{\theta}]}$ where $\mathcal{F}$ denotes the Fisher information matrix.} in one dimension but not in general.  The equivalence is obtained by invoking the \textit{Bernstein Von Mises} theorem: setting $p(\boldsymbol{\theta} | \mathcal{D}) \approx \text{N}(\boldsymbol{\theta}_{\text{MLE}}, \mathcal{F}^{-1}(\boldsymbol{\theta}))$ where $\boldsymbol{\theta}_{\text{MLE}}$ is the maximum likelihood estimate and $\mathcal{F}$ is the Fisher information matrix.  RPs, also like the Jeffreys, are invariant to model reparametrization, which follows from the fact that the mutual information is itself invariant to a change in parametrization \cite{berger2009formal}.

\subsection{RELATED WORK}
Next we  review existing techniques for approximating intractable RPs.  These methods have a notable lack of scalability, requiring numerical integration over the parameter space.  Nonetheless, since they share some fundamental similarities with our proposed method, we reproduce their main components so that we can later discuss how our method handles the same analytical difficulties. 

\textbf{Numerical Algorithm.}  Berger et al.\ (2009) proposed a numerical method for computing a RP's value at any given point $\boldsymbol{\theta}_{0}$.  Their method is, simply, to calculate $f(\boldsymbol{\theta})$ numerically via Monte Carlo approximations: \begin{equation}\label{explicit}
p^{*}(\boldsymbol{\theta}_{0}) \approx \exp \left \{ \frac{1}{J} \sum_{j=1}^{J}\ \log \frac{p(\hat{\mathcal{D}}_{j} | \boldsymbol{\theta}_{0}) \mathds{1}_{\boldsymbol{\Theta}}}{\sum_{s=1}^{S} p(\hat{\mathcal{D}}_{j} | \hat{\boldsymbol{\theta}}_{s})} \right  \}
\end{equation} where $\mathds{1}_{\boldsymbol{\Theta}}$ is an improper uniform prior over the parameter space.  The method proceeds by sampling $J$ datasets from the likelihood function, i.e. $ \hat{\mathcal{D}}_{j} = \{ \hat{\mathbf{x}}_{j,i} | \hat{\mathbf{x}} \sim p(\mathbf{x}|\boldsymbol{\theta}_{0})\}$, and $S$ parameter values from the prior, i.e. $\hat{\boldsymbol{\theta}}_{s} \sim \mathds{1}_{\boldsymbol{\Theta}}$.  The posterior is then approximated as $p(\boldsymbol{\theta} | \mathcal{D}) \approx p(\hat{\mathcal{D}}_{j} | \boldsymbol{\theta}_{0})\mathds{1}_{\boldsymbol{\Theta}} / \sum_{s=1}^{S} p(\hat{\mathcal{D}}_{j} | \hat{\boldsymbol{\theta}}_{s})$.  This numerical approximation has two significant downsides.  The first is that the  {user} must specify the points at which to compute the prior, and the second is that numerically integrating over the parameter space is computationally expensive in even low dimensions.   

\textbf{MCMC.}  Lafferty \& Wasserman (2001) proposed a Markov Chain Monte Carlo (MCMC) method for sampling from a RP.  Their approach involves running the Metropolis-Hastings algorithm on the following ratio \cite{lafferty2001iterative}: \begin{equation}\label{mcmc}\begin{split}
    \log \frac{p^{t+1}(\boldsymbol{\theta})}{p^{t+1}(\boldsymbol{\theta'})} = (t&+1)(\mathbb{H}_{\mathbf{x}|\boldsymbol{\theta'}}[\mathbf{x}] - \mathbb{H}_{\mathbf{x}|\boldsymbol{\theta}}[\mathbf{x}]) + \\ &\sum_{\mathbf{x} \in \mathcal{X}} W^{t}(\mathbf{x})[p(\mathbf{x} | \boldsymbol{\theta'}) -  p(\mathbf{x} | \boldsymbol{\theta})]    
\end{split}\end{equation} where $t$ is the iteration index,
$\mathbb{H}_{\mathbf{x}|\boldsymbol{\theta}}[\mathbf{x}]$ is the entropy of the likelihood function, and $W^{t}(\mathbf{x}) = W^{t-1}(\mathbf{x}) + \log \frac{1}{S_{t}}\sum_{s=1}^{S_{t}} p(\boldsymbol{x} | \hat{\boldsymbol{\theta}}^{t}_{s})$ where $\hat{\boldsymbol{\theta}}^{t}_{s}$ are the parameter samples collected during the previous iteration.  

While this MCMC approach may look dissimilar to Berger et al.'s method at first glance, the two methods are in fact related.  We can see the connection by examining just one of the distributions in the ratio (when $t=0$): \begin{equation*}\label{mcmcConnect}\begin{split}
  \log& \ p^{1}(\boldsymbol{\theta}) = {-\mathbb{H}_{\mathbf{x}|\boldsymbol{\theta}}[\mathbf{x}]} + \sum_{\mathbf{x} \in \mathcal{X}} {-W^{0}(\mathbf{x})} p(\mathbf{x} | \boldsymbol{\theta}) \\ &=  \sum_{\mathbf{x} \in \mathcal{X}} p(\mathbf{x} | \boldsymbol{\theta}) \left[\log p(\mathbf{x} | \boldsymbol{\theta}) -  \log \frac{1}{S_{0}}\sum_{s=1}^{S_{0}} p(\boldsymbol{x} | \hat{\boldsymbol{\theta}}^{0}_{s}) \right].
\end{split}
\end{equation*}   
The line above becomes equivalent to Equation \ref{explicit} if we use a Monte Carlo approximation of the expectation over $p(\mathbf{x}|\boldsymbol{\theta})$ and then exponentiate both sides.  Despite this close connection between the two methods, Lafferty \& Wasserman (2001)'s approach is superior to that of Berger et al.'s since it draws samples from the prior instead of merely computing its value at points the user must select.  Yet the same costly discrete approximations of the integrals will be required.

\textbf{Reference Distance Method.}  The third approach, and the only other that we are aware of for finding approximate RPs, is the \textit{Reference Distance Method} (RDM) proposed by Berger et al.\ (2015).  This method focuses on finding a joint RP by minimizing the divergence between a parametric family and the marginal RPs \cite{berger2015overall}.  Since we are concerned with models for which even the marginal RPs are intractable, the RDM is not a relevant point for comparison.  

\section{LEARNING REFERENCE PRIOR APPROXIMATIONS}
We now turn to the primary contribution of this paper: approximating RPs by \emph{learning} the parameters of the approximation.  Our proposed approach contrasts with Berger et al.\ (2009)'s and Lafferty \& Wasserman (2001)'s in that their methods are not model-based. In other words, their procedures produce no parametric artifact for the prior unless a post-hoc step of model fitting is carried out.  Our black-box optimization framework subsumes the utility of the numerical and MCMC methods as it can directly learn either a parametric approximation to evaluate the prior's density or a functional sampler that can generate new samples from the prior at any later time.    

\subsection{METHOD \#1: INFORMATION LOWER BOUND}\label{m1}
Inspired by recent advances in posterior variational inference (VI), we use similar ideas to optimize an approximate \textit{prior}---call it $p_{\boldsymbol{\lambda}}(\boldsymbol{\theta})$ with parameters $\boldsymbol{\lambda}$---so that it is the distribution in the family closest to the true RP $p^{*}(\boldsymbol{\theta})$.  The mutual information still serves as the natural optimization objective; the difference is  that we take the $\argmax$ over $\boldsymbol{\lambda}$, instead of the density $p$ itself, such that $p^{*}(\boldsymbol{\theta}) \approx p_{\boldsymbol{\lambda}^{*}}(\boldsymbol{\theta})$: 
\begin{equation}\label{refPriorVI}\begin{split}
    &\boldsymbol{\lambda}^{*} = \argmax_{\boldsymbol{\lambda}} \ \  I(\boldsymbol{\theta}, \mathcal{D}) \\ &=  \argmax_{\boldsymbol{\lambda}}  \int_{\boldsymbol{\theta}} p_{\boldsymbol{\lambda}}(\boldsymbol{\theta}) \int_{\mathcal{D}} p(\mathcal{D}|\boldsymbol{\theta}) \log \frac{p(\mathcal{D}, \boldsymbol{\theta})}{p_{\boldsymbol{\lambda}}(\boldsymbol{\theta})p(\mathcal{D})} d\mathcal{D} d\boldsymbol{\theta} \\ &=  \argmax_{\boldsymbol{\lambda}}  \int_{\boldsymbol{\theta}} p_{\boldsymbol{\lambda}}(\boldsymbol{\theta}) \int_{\mathcal{D}} p(\mathcal{D}|\boldsymbol{\theta}) \log \frac{p(\mathcal{D}| \boldsymbol{\theta})}{p(\mathcal{D})}  d\mathcal{D} d\boldsymbol{\theta} \\ &= \argmax_{\boldsymbol{\lambda}} \ \mathbb{E}_ {\boldsymbol{\theta}_{\boldsymbol{\lambda}}} \left[ -\mathbb{H}_{\mathcal{D}|\boldsymbol{\theta}}[\mathcal{D}] - \mathbb{E}_{\mathcal{D}|\boldsymbol{\theta}} [ \log p(\mathcal{D})] \right] . 
\end{split}
\end{equation} 
In the final line above, we wrote the mutual information as the difference between the negative likelihood entropy and the expected log marginal likelihood because this is $I(\boldsymbol{\theta}, \mathcal{D})$'s most tractable form: it contains only $p(\mathcal{D})$ instead of $p(\mathcal{D})$ \emph{and} $p(\boldsymbol{\theta}|\mathcal{D})$.  We use the notional $\boldsymbol{\theta}_{\boldsymbol{\lambda}}$ to emphasize that $\boldsymbol{\theta}$'s distribution is a function of $\boldsymbol{\lambda}$.   

\textbf{Bounding log p($\bm{\mathcal{D}}$).}  The marginal likelihood term in Equation \ref{refPriorVI} is still problematic, and thus, just as in posterior VI, we need some tractable bound to optimize instead.  Since we need to bound $I(\boldsymbol{\theta}, \mathcal{D}) $ from below, $\log p(\mathcal{D})$ must be bounded from \emph{above}.  Hence, unfortunately, we cannot use the \textit{evidence lower bound} (ELBO) common in posterior VI.  As an alternative we use the \textit{variational R\'enyi bound} \cite{li2016variational} (VR), which is defined as: $$\log p(\mathcal{D}) \le \frac{1}{1-\alpha}\log \mathbb{E}_{\boldsymbol{\theta}} \left[ p(\mathcal{D}|\boldsymbol{\theta})^{1-\alpha} \right] \ \text{ for } \ \alpha \le 0.$$  Plugging the VR bound into Equation \ref{refPriorVI} yields a general lower bound on the mutual information: \begin{equation}\label{boundedI}\begin{split}
    &I(\boldsymbol{\theta}, \mathcal{D}) \\ &\ge \mathbb{E}_ {\boldsymbol{\theta}_{\boldsymbol{\lambda}}} \left[ -\mathbb{H}_{\mathcal{D}|\boldsymbol{\theta}}[\mathcal{D}] - \frac{1}{1-\alpha}\log \mathbb{E}_{\boldsymbol{\theta}} \left[ p(\mathcal{D}|\boldsymbol{\theta})^{1-\alpha} \right] \right]. 
\end{split}
\end{equation}  In theory, setting $\alpha=0$ provides the tightest bound, and decreasing $\alpha$ loosens the bound.  However, as we discuss next, practical implementation requires a negative value for $\alpha$.

\textbf{Optimization Objective.}  The expectation within the VR bound usually will not be analytically solvable, requiring the use of a Monte Carlo approximation (which we will refer to as MC-VR).  Yet, introducing sampling into the VR bound can give rise to numerical challenges.  The MC-VR estimator is an exponentiated form of the \textit{harmonic mean estimator} \cite{raftery2006estimating}, which is notorious for its high variance. Furthermore, approximating the expectation with samples, since they reside inside the logarithm, biases the bound downward.  Li \& Turner (2016) propose the following \textit{VR-max} estimator, corresponding to $\alpha \rightarrow -\infty$, to cope with these issues: $\max_{s} \log p(\mathcal{D}|\hat{\boldsymbol{\theta}}_{s})$ where $s$ indexes samples $\hat{\theta}_{s} \sim p(\boldsymbol{\theta})$.  We find that the VR-max estimator generally preserves the bound and needs to be checked only in high dimensions ($100+$), which is a regime not well suited for reference priors anyway (due to overfitting).  

Introducing the VR-max estimator into Equation \ref{refPriorVI} yields a tractable lower bound on the mutual information: \begin{equation}\label{J}\begin{split}
& I(\boldsymbol{\theta}, \mathcal{D}) \ge \mathcal{J}_{\text{RP}}(\boldsymbol{\lambda}) \\ &= \mathbb{E}_ {\boldsymbol{\theta}_{\boldsymbol{\lambda}}} \left[ {-\mathbb{H}_{\mathcal{D}|\boldsymbol{\theta}}[\mathcal{D}]} - \mathbb{E}_{\mathcal{D}|\boldsymbol{\theta}} [\max_{s} \log p(\mathcal{D}|\hat{\boldsymbol{\theta}}_{s})]  \right].\end{split}
\end{equation}  Maximizing $\mathcal{J}_{\text{RP}}(\boldsymbol{\lambda})$ with respect to the prior's parameters $\boldsymbol{\lambda}$ results in $p_{\boldsymbol{\lambda}}(\boldsymbol{\theta}) \approx p^{*}(\boldsymbol{\theta})$ as long as $p_{\boldsymbol{\lambda}}$ is sufficiently expressive.  $\mathcal{J}_{\text{RP}}(\boldsymbol{\lambda})$ can be interpreted as follows.  The first term is the entropy of the likelihood function, and thus maximizing its negation encourages certainty in the data model.  The second term, the expected value of the VR-max estimator under the likelihood, encourages diversity in $p_{\boldsymbol{\lambda}}$ by forcing a dataset $\mathcal{D}_{0} \sim p(\mathcal{D} | \boldsymbol{\theta}_{0})$ to have low probability under other parameter settings $\hat{\boldsymbol{\theta}}_{s}$. 

\textbf{Connection to Previous Work.}  Further understanding of $\mathcal{J}_{\text{RP}}(\boldsymbol{\lambda})$ can be gained by re-writing it to see its relationship to Berger et al.\ (2009)'s and Lafferty \& Wasserman (2001)'s methods.  Pulling out the expectation over the likelihood, we have the equivalent form: $$ \mathcal{J}_{\text{RP}}(\boldsymbol{\lambda}) = \mathbb{E}_ {\boldsymbol{\theta}_{\boldsymbol{\lambda}}} \mathbb{E}_ {\mathcal{D} | \boldsymbol{\theta}} \left[ \log p(\mathcal{D} | \boldsymbol{\theta}) - \max_{s} \log p(\mathcal{D}|\hat{\boldsymbol{\theta}}_{s})  \right], $$ which is the difference between the data's log-likelihood under the model (i.e. parameter setting) that generated this data and the data log-likelihood under several samples from the prior. 
We see that optimization forces the prior to place most of its mass on parameters that generate identifiable datasets---or in other words, datasets that have high probability under only their true generative model.  Turning back to Berger et al.\ (2009)'s Equation \ref{explicit}, and recalling its connection to the MCMC method, we see each method is approximately computing $\log [p(\mathcal{D} | \boldsymbol{\theta}) / p(\mathcal{D})]$ with the critical difference being that Berger et al.\ (2009) and Lafferty \& Wasserman (2001) approximate $\log p(\mathcal{D})$ with $\log \frac{1}{S} \sum_{s} p(\mathcal{D}|\hat{\boldsymbol{\theta}}_{s}) $ whereas we use $\max_{s} \log p(\mathcal{D}|\hat{\boldsymbol{\theta}}_{s})$ in order to ensure a proper lower bound. 

\subsubsection{Black-Box, Gradient-Based Optimization}  We now address how to compute and optimize $\mathcal{J}_{\text{RP}}(\boldsymbol{\lambda})$ (Equation \ref{J}) efficiently using differentiable Monte Carlo approximations.

\textbf{Computing the Expectations.}  Consider first the three expectations in Equation \ref{J}. Starting with the $\mathbb{H}_{\mathcal{D}|\boldsymbol{\theta}}[\mathcal{D}]$ term, for many predictive models, $p(\mathbf{x}|\boldsymbol{\theta})$
is either Gaussian, as in regression, or Bernoulli, as in binary classification, meaning $\mathbb{H}_{\mathcal{D}|\boldsymbol{\theta}}[\mathcal{D}]$ can be computed analytically\footnote{To keep the notation simple, in our discussion of conditional models the dependence on the features is implicit.  Writing the entropy with $\mathbf{X'}$ as the feature matrix and $\mathbf{x}$ as the vector of labels, we have: $\mathbb{H}_{\mathbf{x}|\mathbf{X'}, \boldsymbol{\theta}}[\mathbf{x}]$.}.  The second term, $\mathbb{E}_{\mathcal{D}|\boldsymbol{\theta}} [\max_{s} \log p(\mathcal{D}|\hat{\boldsymbol{\theta}}_{s})]$, is simply the cross-entropy between $p(\mathcal{D}|\boldsymbol{\theta})$ and $p(\mathcal{D}|\hat{\boldsymbol{\theta}}_{\text{max}})$ where $\hat{\boldsymbol{\theta}}_{\text{max}}$ is the sample that maximizes the likelihood.  This term also can  usually be calculated analytically for regression and classification models.  The only component that will typically be intractable is the expectation over $p_{\boldsymbol{\lambda}}(\boldsymbol{\theta})$, as the parameters are often buried under nonlinear functions and nested hierarchies.  To address this we compute the outer expectation with samples $\hat{\boldsymbol{\theta}} \sim p_{\boldsymbol{\lambda}}(\boldsymbol{\theta})$: \begin{equation}\begin{split}\label{mcJ}
    \tilde{\mathcal{J}}_{\text{RP}}(\boldsymbol{\lambda}) &= \frac{1}{S} \sum_{s=1}^{S} \mathbb{H}[p(\mathcal{D}|\hat{\boldsymbol{\theta}}_{s}) || p(\mathcal{D}|\hat{\boldsymbol{\theta}}_{\text{max}}) ] - \mathbb{H}_{\mathcal{D}|\hat{\boldsymbol{\theta}}_{s}}[\mathcal{D}] \\ &= \frac{1}{S} \sum_{s=1}^{S} \text{KLD}[ p(\mathcal{D}|\hat{\boldsymbol{\theta}}_{s}) \mid\mid p(\mathcal{D}|\hat{\boldsymbol{\theta}}_{\text{max}}) ]
\end{split}\end{equation} for $S$ samples from the RP approximation and where $\mathbb{H}[p(\mathcal{D}|\hat{\boldsymbol{\theta}}_{s}) || p(\mathcal{D}|\hat{\boldsymbol{\theta}}_{\text{max}}) ]$ denotes the cross-entropy term mentioned above.  If both entropy terms can be computed analytically, we can write the expression as a KLD, which we do in the second line by using the identity $\text{KLD}[q||p] = \mathbb{H}[q||p]-\mathbb{H}[q]$.  If the entropy terms are not analytically tractable, they will need to be estimated by sampling from the likelihood function.  

\textbf{Differentiable Sampling:}  
We can take derivatives through each $\hat{\boldsymbol{\theta}}_{s}$, thereby allowing for fully gradient-based optimization, by drawing the samples via a differentiable non-centered parametrization (DNCP)---the so-called `reparametrization trick' \cite{kingma2013auto}, i.e. \begin{equation*}\begin{split} \frac{\partial}{\partial \boldsymbol{\lambda}} &\left[ \text{KLD}[ p(\mathcal{D}|\hat{\boldsymbol{\theta}}_{s}) \mid\mid p(\mathcal{D}|\hat{\boldsymbol{\theta}}_{\text{max}}) ] \right] = \\ & \frac{\partial}{\partial \hat{\boldsymbol{\theta}}}\left[ \text{KLD}[ p(\mathcal{D}|\hat{\boldsymbol{\theta}}_{s}) \mid\mid p(\mathcal{D}|\hat{\boldsymbol{\theta}}_{\text{max}}) ] \right] \frac{\partial \hat{\boldsymbol{\theta}}}{\partial \boldsymbol{\lambda}} \end{split}\end{equation*} where $\frac{\partial \hat{\boldsymbol{\theta}}}{\partial \boldsymbol{\lambda}}$ is the derivative that needs a DNCP in order to be evaluated.  Requiring that $p_{\lambda}$ has a DNCP does not significantly limit the approximating family.  For instance, most mixture densities have a DNCP.  When dealing with discrete data or parameters, we can use the \textit{Concrete distribution} \cite{2016arXiv161100712M, jang2016}, a differentiable relaxation of the discrete distribution, to still have fully gradient-based learning. 

\subsubsection{Implicit Priors}\label{ips}
A crucial detail to note about Equation \ref{mcJ} is that it does not require evaluation of the prior's density.  Rather, we need only to draw samples from it.  This allows us to use black-box functional samplers as the variational family \cite{ranganath2016operator}, i.e. $\hat{\boldsymbol{\theta}} = g(\boldsymbol{\lambda}, \hat{\boldsymbol{\epsilon}}) \text{ where } \boldsymbol{\epsilon} \sim p_{0} $, $g$ is some arbitrary differentiable function (such as a neural network), and $p_{0}$ is a fixed noise distribution.  We call $p_{\boldsymbol{\lambda}}$ an \textit{implicit prior} in this setting since its density function is unknown.  

Thus, the proposed information bound provides a `built-in' sampling technique in lieu of Lafferty \& Wasserman (2001)'s MCMC algorithm.  Although we cannot guarantee the same asymptotically unbiased approximation as MCMC, the lack of restrictions on $g(\boldsymbol{\lambda}, \hat{\boldsymbol{\epsilon}})$ should allow for a sufficiently expressive sampler.  Furthermore, we can persist the sampler just by saving the values of $\boldsymbol{\lambda}$; there's no need to save the samples themselves.  And since learning a RP for a generative model is dataset  {independent}, $\boldsymbol{\lambda}$ could be shared easily via an online repository and users desiring a RP for the same model could download $\boldsymbol{\lambda}$ to generate an unbounded number of their own samples.  The same can be done when $p_{\boldsymbol{\lambda}}$ is a proper distribution.

\subsubsection{Example: Gaussian Mean}
To provide some intuition and to sanity check the proposed approach, consider learning an approximate RP for the mean parameter $\boldsymbol{\mu}$ of a Gaussian density.
The RP on $\boldsymbol{\mu}$ is the improper uniform distribution, which can be approximated as a Gaussian with infinite variance: $p^{*}(\boldsymbol{\mu}) \propto \mathds{1} \approx \text{N}(\cdot, \infty).$  The analytical solution to the KLD term in Equation \ref{mcJ} in this case is: $$  \text{KLD}[ p(\mathcal{D}|\hat{\boldsymbol{\theta}}_{s}) \mid\mid p(\mathcal{D}|\hat{\boldsymbol{\theta}}_{\text{max}}) = \frac{1}{2}\mid\mid \hat{\boldsymbol{\mu}}_{s} - \hat{\boldsymbol{\mu}}_{\text{max}}  \mid\mid_{2}^{2},$$ which is the squared distance between two samples from $p_{\boldsymbol{\lambda}}$.  Maximizing Equation \ref{mcJ} therefore maximizes the average distance between samples from the RP approximation.  If we set $p_{\boldsymbol{\lambda}} = \text{N}(\boldsymbol{\mu}_{\boldsymbol{\lambda}}, \sigma^{2}_{\boldsymbol{\lambda}}\mathbb{I})$ and transform to the Normal's DNCP $\boldsymbol{\theta} = \boldsymbol{\mu}+\boldsymbol{\sigma}\odot \boldsymbol{\epsilon}$ where $\boldsymbol{\epsilon}\sim \text{N}(\mathbf{0}, \mathbb{I})$, then the optimization objective becomes $$ \mid\mid \hat{\boldsymbol{\mu}}_{s} - \hat{\boldsymbol{\mu}}_{\text{max}}  \mid\mid_{2}^{2} \ \ = \ \ \mid\mid \boldsymbol{\sigma}_{\boldsymbol{\lambda}} \odot (\hat{\boldsymbol{\epsilon}}_{s} - \hat{\boldsymbol{\epsilon}}_{\text{max}})  \mid\mid_{2}^{2}, $$ and  optimization would increase $\sigma^{2}_{\boldsymbol{\lambda}}$ without bound, agreeing with the infinite-variance Normal approximation.

\subsection{METHOD \#2: PARTICLE DESCENT}\label{m2}
Next we present a particle-based approximation method, which we outline below.  While the core ideas are not significantly different from those of Method \#1, here we use a different rearrangement of the mutual information and a lower bound bound on $\log p(\mathcal{D})$.  These changes expose nuances that may be useful if the reader wishes to apply other VI techniques to the problem of RP approximation.  Other modern VI methods using transformations \cite{rezende2015variational} or importance weighting \cite{burda2015importance} could also be applied.

\textbf{A Particle-Based Approximation.}  \textit{Stein Variational Gradient Descent} \cite{liu2016stein} (SVGD) is a variational inference technique that exploits a connection between the Stein operator and the derivative of the KLD to derive a deterministic particle update rule.  At time step $t$, each parameter particle $\bar{\boldsymbol{\theta}}_{j}$ is updated according to: \begin{equation*}\begin{split}
&\bar{\boldsymbol{\theta}}_{j}^{t+1} = \bar{\boldsymbol{\theta}}_{j}^{t} + \eta \  \phi[{\boldsymbol{\theta}}]  \ \ \ \text{ where } \\ &\phi[{\boldsymbol{\theta}}] = \frac{1}{K} \sum_{k=1}^{K} \kappa(\bar{\boldsymbol{\theta}}_{k}^{t}, \bar{\boldsymbol{\theta}}_{j}^{t}) \nabla_{\bar{\boldsymbol{\theta}}_{k}} \log p(\bar{\boldsymbol{\theta}}_{k}^{t}) + \nabla_{\bar{\boldsymbol{\theta}}_{k}} \kappa(\bar{\boldsymbol{\theta}}_{k}^{t}, \bar{\boldsymbol{\theta}}_{j}^{t}).
\end{split}\end{equation*}  $\eta$ is a learning rate, $\kappa(\cdot, \cdot)$ is a proper kernel function, and $p$ is the cumbersome distribution we wish to approximate.  SVGD has the nice property that using one particle reduces to vanilla gradient ascent on $p$ (MAP estimation, in the Bayesian setting).     

Similarly to $\tilde{\mathcal{J}}_{\text{RP}}(\boldsymbol{\lambda})$ (Equation \ref{mcJ}), SVGD does not require the approximating density be evaluated, and hence we can draw the particles from a black-box function, i.e. $\bar{\boldsymbol{\theta}} = g(\boldsymbol{\lambda}, \hat{\boldsymbol{\epsilon}})$, just as defined in Section \ref{ips}.  Liu \& Feng (2016) call this variant \textit{Amortized SVGD} (A-SVGD), and it estimates the parameters $\boldsymbol{\lambda}$ by finding SVDG's fixed-point solutions \cite{liu2016wild}: \begin{equation}\label{Jasvgd}\begin{split}
    \hat{\boldsymbol{\lambda}} &=  \argmin_{\boldsymbol{\lambda}} \mid\mid g(\boldsymbol{\lambda}, \hat{\boldsymbol{\epsilon}}) - (\bar{\boldsymbol{\theta}} + \eta \  \phi[{\boldsymbol{\theta}}] ) \mid\mid^{2}_{2} \\ & = \argmin_{\boldsymbol{\lambda}} \mid\mid \eta \ \phi[{\boldsymbol{\theta}}] \mid\mid^{2}_{2}.\end{split}\end{equation} 
This formulation is especially beneficial to SVGD because training can be done with a practical number of particles ($K$), but an unlimited number can be drawn at evaluation time.  Lastly, note that the base parameters are updated through the $\bar{\boldsymbol{\theta}}_{j}^{t}$ particle, not through the particle in $\nabla_{\bar{\boldsymbol{\theta}}_{k}^{t}} \log p(\bar{\boldsymbol{\theta}}_{k}^{t})$.     

\textbf{A-SVGD for RP Approximations.}  We can use A-SVGD to learn particle approximations of RPs.  A-SVGD requires RP learning be formulated as KLD minimization. In this context recall Equation \ref{refPrior3}  in which we showed  that minimizing $\text{KLD}[p(\boldsymbol{\theta}) \mid \mid f(\boldsymbol{\theta})]$\footnote{While Equation \ref{refPrior3}, in form, looks like a negated KLD, it will take on positive values due to $f$ being unnormalized.} maximizes $I(\boldsymbol{\theta}, \mathcal{D})$.  Hence we can treat the functional $f$ as the intractable density on which to apply the SVGD operator $\phi[\boldsymbol{\theta}]$.  The gradient term is then \begin{equation}\begin{split}\label{asvgd}
& \nabla_{\bar{\boldsymbol{\theta}}} \log  f(\bar{\boldsymbol{\theta}}) = \nabla_{\bar{\boldsymbol{\theta}}} \int_{\mathcal{D}} p(\mathcal{D}|\bar{\boldsymbol{\theta}}) \ \log p(\bar{\boldsymbol{\theta}} | \mathcal{D}) \ d \mathcal{D}  \\ &= \nabla_{\bar{\boldsymbol{\theta}}} \int_{\mathcal{D}} p(\mathcal{D}|\bar{\boldsymbol{\theta}}) \log \frac{p(\mathcal{D}|\bar{\boldsymbol{\theta}}) \mathds{1}_{\boldsymbol{\Theta}}}{p(\mathcal{D})} d \mathcal{D} \\ &=  {-} \nabla_{\bar{\boldsymbol{\theta}}}  \mathbb{H}_{\mathcal{D}|\bar{\boldsymbol{\theta}}}[\mathcal{D}]  - \nabla_{\bar{\boldsymbol{\theta}}}  \mathbb{E}_{\mathcal{D}|\bar{\boldsymbol{\theta}}}[\log p(\mathcal{D})]    \end{split}
\end{equation} where $\bar{\boldsymbol{\theta}}$ is the particle.  In line 2, following Berger et al.\ (2009) in Equation \ref{explicit}, we assume the prior is constant.  If a prior is included, it acts as a hyper-prior, regularizing the particles towards a user-specified distribution.  

Again we face the problem of evaluating $\log p(\mathcal{D})$.  We could use the VR-max estimator, but here we opt for stability and use the ELBO: $\log p(\mathcal{D}) \ge \mathbb{E}_{\boldsymbol{\theta}_{\boldsymbol{\lambda}}}[\log p(\mathcal{D} | \boldsymbol{\theta})]$.  Making this substitution reduces the influence of $\partial p(\mathcal{D}|\bar{\boldsymbol{\theta}}) / \partial \bar{\boldsymbol{\theta}}$, the term that encourages diversity in $p_{\boldsymbol{\lambda}}$ and the generation of identifiable datasets (as discussed in Section \ref{m1}).  Yet the derivative of the kernel, the second term in $\phi[\boldsymbol{\theta}]$, acts as a (locally) repulsive force on the particles, which may compensate for the introduction of the ELBO.  Substituting the approximation of the marginal likelihood into Equation \ref{asvgd} yields:
\begin{equation}\begin{split} &\approx {-}\nabla_{\bar{\boldsymbol{\theta}}} \ \mathbb{H}_{\mathcal{D}|\bar{\boldsymbol{\theta}}}[\mathcal{D}]  - \nabla_{\bar{\boldsymbol{\theta}}} \  \mathbb{E}_{\mathcal{D}|\bar{\boldsymbol{\theta}}} \ \mathbb{E}_{\boldsymbol{\theta}_{\boldsymbol{\lambda}}}[\log p(\mathcal{D} | \boldsymbol{\theta})]
\\ &= {-}\nabla_{\bar{\boldsymbol{\theta}}} \ \mathbb{H}_{\mathcal{D}|\bar{\boldsymbol{\theta}}}[\mathcal{D}] + \nabla_{\bar{\boldsymbol{\theta}}} \  \mathbb{E}_{\boldsymbol{\theta}_{\boldsymbol{\lambda}}} \  \mathbb{H}[p(\mathcal{D}|\bar{\boldsymbol{\theta}}) || p(\mathcal{D}| \boldsymbol{\theta}) ] \\ &\approx  \nabla_{\bar{\boldsymbol{\theta}}} \  \frac{1}{S} \sum_{s} \mathbb{H}[p(\mathcal{D}|\bar{\boldsymbol{\theta}}) || p(\mathcal{D}| \hat{\boldsymbol{\theta}}_{s}) ] - \nabla_{\bar{\boldsymbol{\theta}}} \ \mathbb{H}_{\mathcal{D}|\bar{\boldsymbol{\theta}}}[\mathcal{D}] \\ &= \nabla_{\bar{\boldsymbol{\theta}}} \  \frac{1}{S} \sum_{s} \text{KLD}[ p(\mathcal{D}|\bar{\boldsymbol{\theta}}) \mid\mid p(\mathcal{D}| \hat{\boldsymbol{\theta}}_{s}) ].
\end{split}
\end{equation} In the last line we use a Monte Carlo approximation of $\mathbb{E}_{\boldsymbol{\theta}_{\boldsymbol{\lambda}}}$ just as  in Method \#1, where $\hat{\boldsymbol{\theta}}_{s}$ denotes a sample from $g$.  Optimization is performed   as before as described in Equation \ref{Jasvgd}. 

\begin{figure}[t]
\centering     
\subfigure[Bernoulli]{\label{fig:a1}\includegraphics[width=25mm]{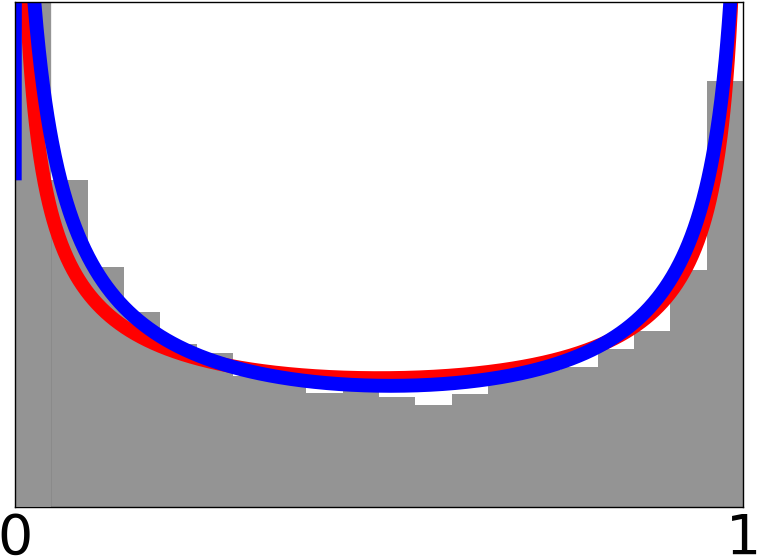}}
\subfigure[Gaussian Scale]{\label{fig:b}\includegraphics[width=25mm]{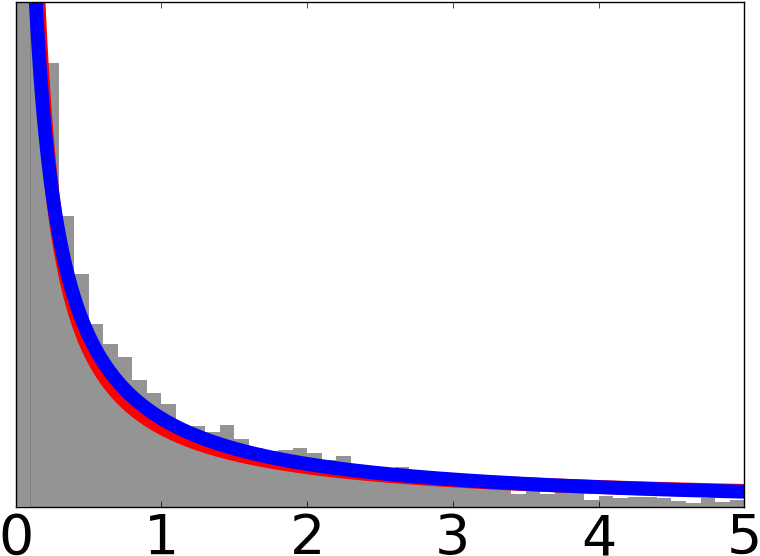}}
\subfigure[Poisson]{\label{fig:c}\includegraphics[width=25mm]{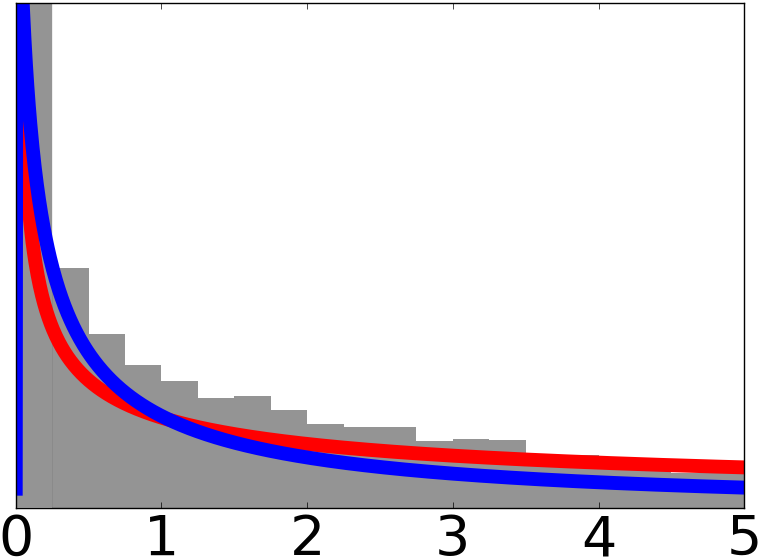}}
\subfigure{\includegraphics[width=40mm]{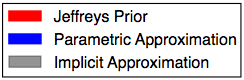}}
\caption{\textit{Approximation via Lower Bound Optimization.}}
\label{qualitative1}
\end{figure}

\section{EMPIRICAL RESULTS}
Below we describe several empirical analyses of the proposed methods.  Formulating experiments is somewhat difficult due to the fact that RPs do not necessarily improve a model's ability to generalize to out-of-sample data.  In fact, using an RP when a model requires regularization will likely degrade performance.  Thus, our main analysis is a case study of the \textit{Variational Autoencoder} \cite{kingma2013auto, rezende2014stochastic}.  But before analyzing the Variational Autoencoder's RP, we check that our methods do indeed recover known RPs for exponential family models.

For all experiments, we used the \textit{AdaM} optimization algorithm \cite{kingma2014adam} with settings $\beta_{1}=0.9$ and $\beta_{2}=0.999$.  Training parameters such as the learning rate, latent dimensionality of the functional sampler $g(\boldsymbol{\lambda}, \hat{\boldsymbol{\epsilon}})$, number of samples were chosen based on which combination gave the highest average value of the information lower bound over the last 50 updates.  The number of training iterations was set at $250$ and the batch size was set to $100$ in all cases. 

\begin{figure}[!ht]
\centering     
\subfigure[Bernoulli]{\label{fig:a}\includegraphics[width=25mm]{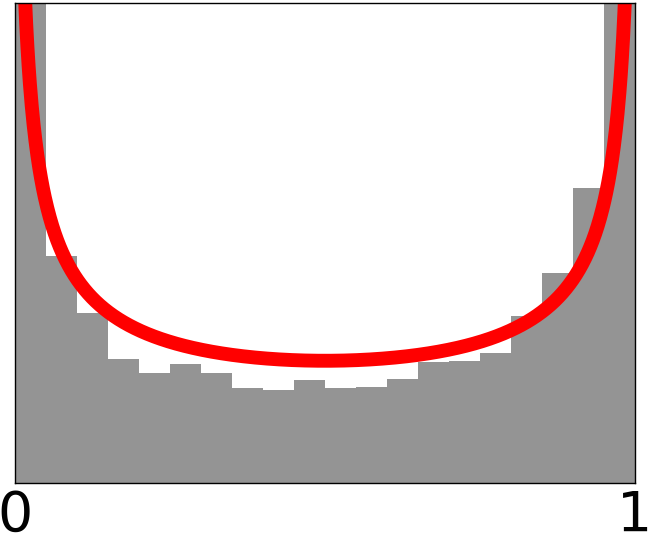}}
\subfigure[Gaussian Scale]{\label{fig:b2}\includegraphics[width=25mm]{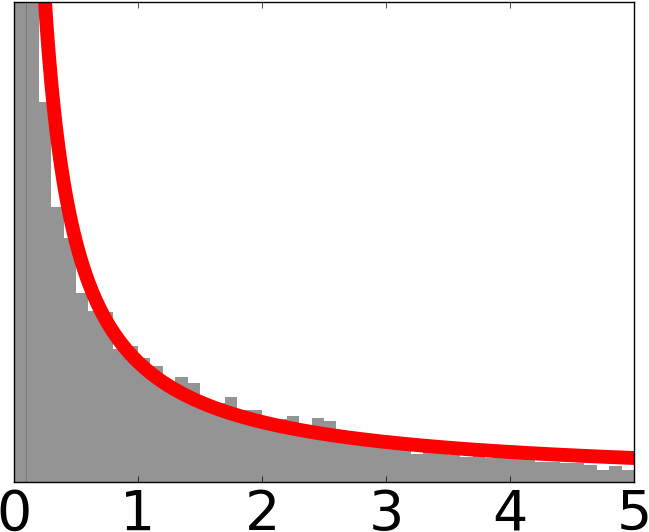}}
\subfigure[Poisson]{\label{fig:c2}\includegraphics[width=25mm]{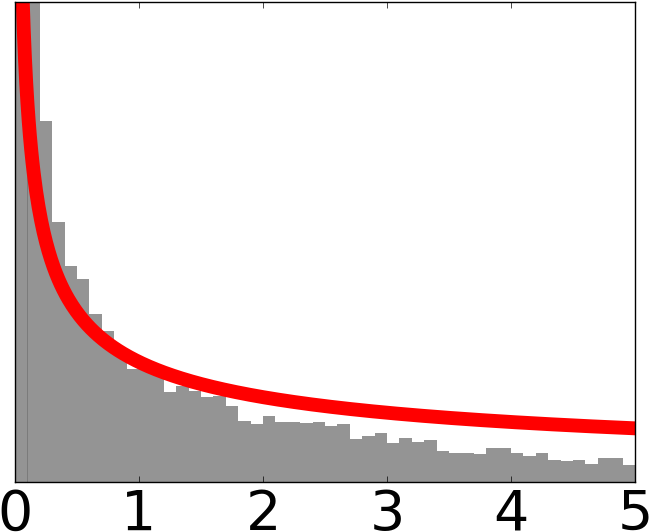}}
\subfigure{\includegraphics[width=39mm]{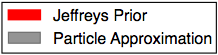}}
\caption{\textit{Approximation via Particle Descent.}}
\label{qualitative2}
\end{figure}

\subsection{RECOVERING JEFFREYS PRIORS}
\begin{figure*}
\centering     
    \subfigure[Bernoulli]{\label{fig:a3}\includegraphics[width=0.25\linewidth]{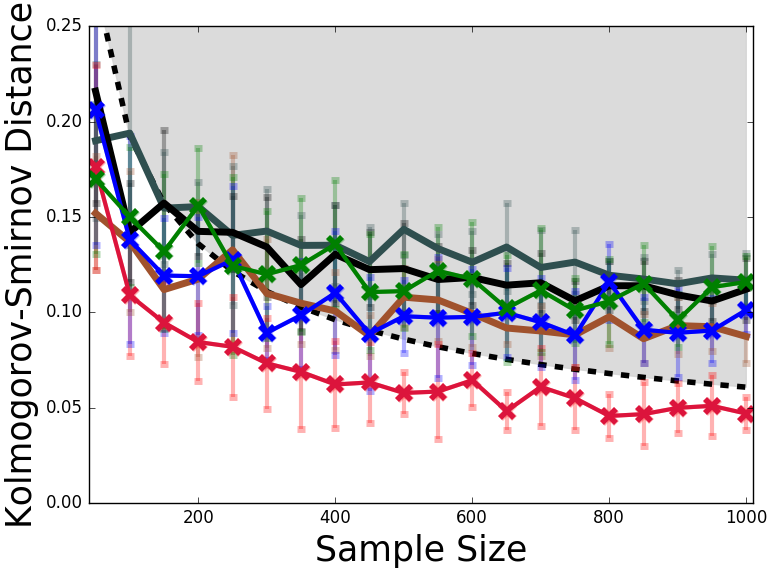}}
    \subfigure[Gaussian Scale]{\label{fig:b3}\includegraphics[width=0.235\linewidth]{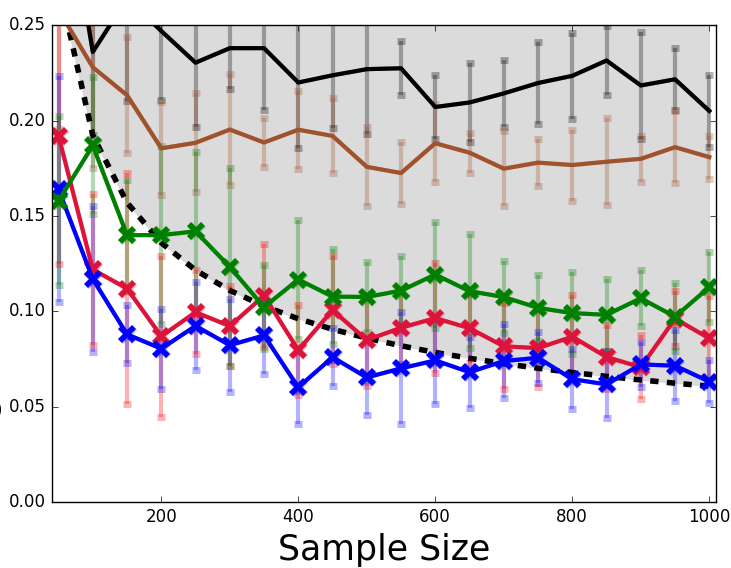}}
    \subfigure[Poisson]{\label{fig:c3}\includegraphics[width=0.235\linewidth]{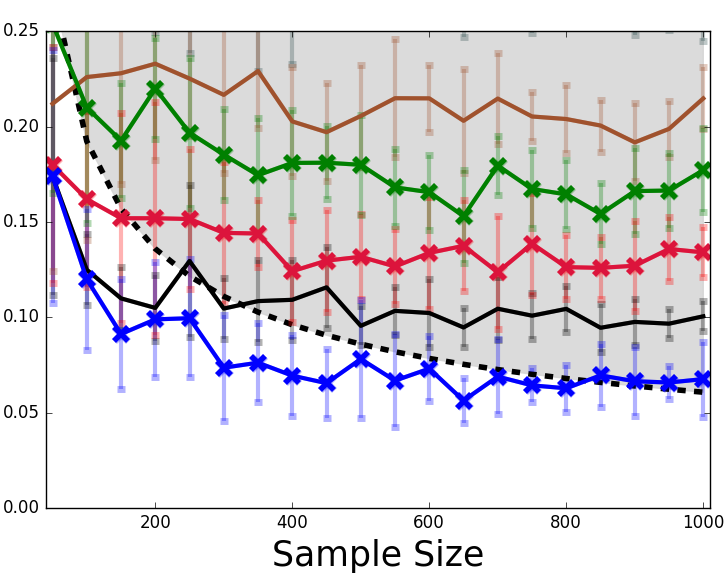}} \hfil
    \subfigure{\includegraphics[width=0.25\linewidth]{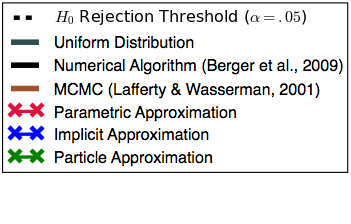}}
\caption{\textit{Quantifying the Approximation Quality.}  The  Kolmogorov-Smirnov distance (supremum of distance between empirical CDFs) between the Jeffreys/true reference prior and the various approximation techniques.  The gray region denotes where the test's null hypothesis is rejected, meaning there is a statistical difference between the distributions.}
\label{expoSims}
\end{figure*}
We begin experimental evaluation by attempting to recover the true RP for three one-dimensional models: the Bernoulli mean parameter, $p^{*}(p) \propto \text{Beta}(.5, .5)$, the Gaussian scale parameter, $p^{*}(\sigma) \propto 1/\sigma$, and the Poisson rate parameter, $p^{*}(\lambda) \propto 1/\sqrt{\lambda}$.  These are also the Jeffreys priors for the respective models (since we are in the univariate case).  The chosen learning rate was $.001$ for the implicit priors and $.0001$ for the parametric and particle approximations, the number of samples drawn was $50$ for all models, and the functional sampler $g$ was a linear model with a latent dimensionality of 5, i.e. $\boldsymbol{\epsilon} \sim \text{N}(\mathbf{0}, \mathbb{I}_{5 \times 5})$.  We used a logit-normal distribution for the Bernoulli RP's parametric approximation and a log-normal for the Gaussian scale's and Poisson's RP approximation.  Both the logit- and log-normal have DNCPs.  For A-SVGD, we used the Sobolev kernel (length scale of $2$) on the unit interval for the Bernoulli model and an RBF in log space (length scale set via the heuristic in \cite{liu2016stein}) for the Gaussian scale and Poisson models. 

\textbf{Qualitative Evaluation.}
Plots of the density functions learned by the lower bound method (Section \ref{m1}) are shown in Figure \ref{qualitative1}.  The red line shows the Jeffreys prior (the gold-standard RP), the blue line shows the parametric approximation, and the gray histogram represents $10,000$ samples from the implicit prior.  Both approximation types have negligible qualitative difference to the red line.

The density functions learned by A-SGVB (Section \ref{m2}) are shown in Figure \ref{qualitative2}.  Again, the red line denotes the true RP, and the gray histogram represents $10,000$ particles sampled from $g$.  Here, we do notice some minor differences.  For instance, the Bernoulli prior's right mode seems to be a bit stronger than its left, and the Poisson prior exhibits underestimation in its tail.  These defects are likely due to the approximations not being penalized as much as in Method \#1 for concentrating their mass in one of the likelihood function's points of low entropy.

\textbf{Quantitative Evaluation.}
Next we quantitatively compare our methods via a two-sample test against three baselines: Berger et al.\ (2009)'s numerical method, Lafferty \& Wasserman (2001)'s MCMC algorithm, and a uniform prior, which serves as a naive flat prior.  For Berger et al.\ (2009)'s method, we use the same number of parameter samples ($S$) as our method, set the $J$ parameter to $100$, and sample datasets containing $500$ points.  To generate samples from Berger et al.\ (2009)'s method, we calculate the prior at $1000$ evenly spaced grid points across the domain and then treat them as a discrete approximation with each point having probability $p(\boldsymbol{\theta}_{i})/\sum_{j}^{1000} p(\boldsymbol{\theta}_{j})$.  We then sample from this discrete distribution $1000$ times.  For the MCMC method, we replicate Lafferty \& Wasserman (2001)'s simulations by using a uniform proposal distribution and running for $10,000$ iterations.  We kept the last $1000$ samples drawn (no need to account for auto-correlation due to the uniform proposal).  For the Gaussian and Poisson cases, we approximated $\mathcal{X}$ using $1000$ points.  For all settings, we made sure our approximation methods ran no longer than the baselines, but this was never an issue: our methods converged in a fraction of the time the numerical algorithms needed to run.  

We quantify the gap in the approximations via a Kolmogorov-Smirnov two-sample test (KST) under the null hypothesis $H_{0}: p=q$ where $p$ is the true RP and $q$ is an approximation.  We draw samples from the true RP, when it is improper, via the same discrete approximation used for Berger et al.\ (2009)'s method.  The KST computes the distance (KSD) between the distributions as $\text{KSD}(p,q) =  \sup_{x} \mid \hat{F}_{p}(x) - \hat{F}_{q}(x) \mid$ where $\hat{F}_{p}(x)$ is the empirical CDF.  

Figure \ref{expoSims} shows the KSD between samples from the Jeffreys prior and the various approximation techniques, as the sample size increases.  The black dotted line in conjunction with the gray shaded area denotes the threshold at which the null hypothesis (that the distributions are equal) is rejected.  The uniform distribution is denoted by the dark gray line, the numerical algorithm by the black, MCMC by the brown, the parametric approximation by the red, the implicit prior by the blue, and the particle approximation by the green.  We see that the latter three approximations (ours) have a lower KSD---and thus are closer to the true RP---in almost every experiment.  The exceptions are that MCMC is superior to A-SVGD for the Bernoulli (and competitive with the implicit), and the Berger et al.\ (2009) technique bests A-SVGD and the parametric approximation for the Poisson.  The parametric approximation for the Bernoulli and the implicit prior for the Gaussian scale and Poisson are the only methods that achieve conspicuous indistinguishability.     

\subsection{OPTIMIZATION STABILITY}
\begin{figure}
\centering     
\subfigure[Samples]{\label{fig:a4}\includegraphics[width=39.25mm]{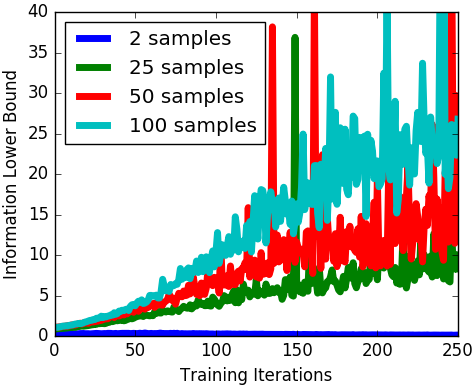}}
\subfigure[Dimensionality]{\label{fig:b4}\includegraphics[width=39.25mm]{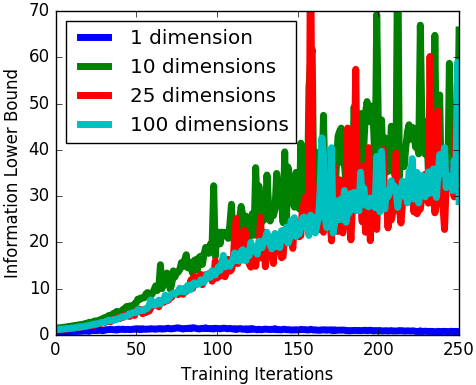}}
\caption{\textit{Optimization Stability.}  We train an implicit prior for a multivariate Gaussian and vary (a) the number of samples used in the VR-max estimator, and (b) the Gaussian's dimensionality.}
\label{trainingAna}
\end{figure}
As discussed in Section \ref{m1}, the VR-max estimator used in Equation \ref{J} has intrinsically high-variance.  While we have just shown in the previous section that our approximations better recover the true RP in one dimension, scaling to higher dimensions is a concern (as is also the case for existing techniques).  Here we examine optimization progress of an RP approximation for the scale parameters of a multivariate Gaussian with a diagonal covariance matrix.  We produce two plots: one showing the information lower bound's progress (for a linear model implicit prior) when using a different number of samples over which to take the maximum (Figure \ref{trainingAna}a) and another showing progress as the Gaussian's dimensionality increases (Figure \ref{trainingAna}b).  For the former, using a five dimensional Gaussian, we see there is a trade-off between lower bound maximization and the number of samples used: using more samples increases the rate of progress but also the objective's variance.  We find that in less than ten dimensions, using around $50$ samples (red line) results in a good variance vs progress balance.  In Figure \ref{trainingAna}b, in which we vary the dimensionality of the Gaussian while keeping the number of samples fixed at $100$, we see that the objective's variance decreases with dimensionality.  While this may seem non-intuitive at first, recall that the VR-max estimator acts as a diversity term, finding points in space that give the data high probability even though it was generated with different parameters.  As dimensionality inflates, it becomes harder and harder for a finite number of samples to capture these points and thus the $-\mathbb{H}$ term in Equation \ref{mcJ} becomes prone to mode seeking. 

\subsection{VAE CASE STUDY}
\begin{figure}
\centering     
\subfigure[Training Configuration]{\label{fig:a5}\includegraphics[width=39.25mm]{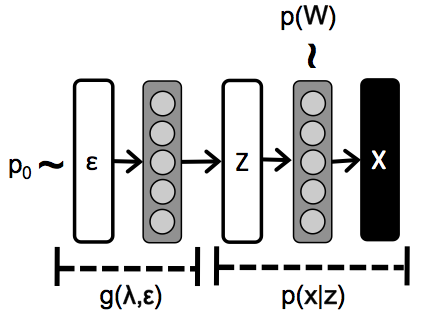}}
\subfigure[Approximation]{\label{fig:b5}\includegraphics[width=39.25mm]{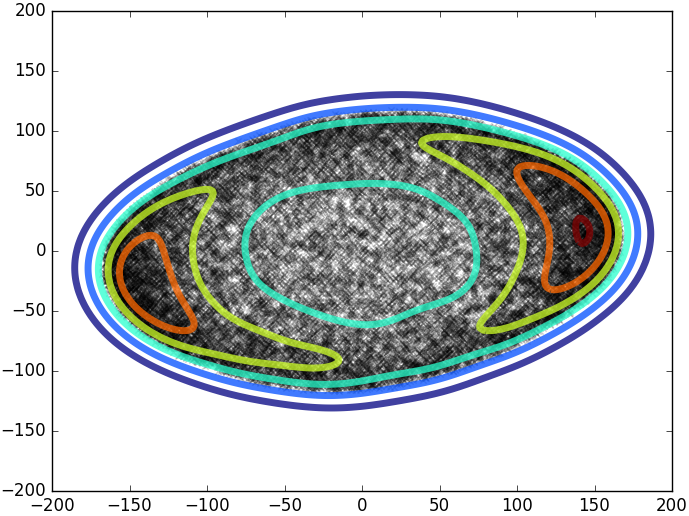}}
\caption{\textit{Learning the Variational Autoencoder's Reference Prior.}  (a)  computational pipeline from the implicit prior through the VAE decoder; (b) RP approximation (contours are generated via kernel density estimation on $10,000$ samples).}
\label{fig:03-03}
\end{figure}
Lastly, we study learning an RP approximation for an intractable, neural-network-based model: a \textit{Variational Autoencoder} \cite{kingma2013auto} (VAE).   The standard Normal distribution is often chosen as the prior on the VAE's latent space \cite{kingma2013auto, rezende2014stochastic, burda2015importance}, and   this choice is made more for analytical simplicity rather than convictions based on prior information\footnote{``I chose the simple Gaussian prior N(0,I) because it's simple to demonstrate but also because it results in a relatively friendly objective function." --- D. Kingma, comment taken from \textit{r/MachineLearning}, 4/12/16.}.  Thus, we learn an RP for the VAE to investigate the qualities of its objective prior, which was previously intractable.

We trained an implicit prior (IP) for a VAE with $784$ output dimensions (MNIST's size), $100$ encoder hidden units with hyperbolic tangent activations, and a two-dimensional latent space for purposes of visualization.  The IP $g(\boldsymbol{\lambda}, \hat{\boldsymbol{\epsilon}})$ is also a one-hidden layer neural network\footnote{Architecture / training paramters: 2000 latent noise dimensions, 1000 hidden dimensions, ReLU activations, $.0003$ learning rate, $50$ samples for VR-max term.}.  The computational pipeline is depicted in Figure \ref{fig:03-03}a, where $p(\mathbf{x} | \mathbf{z} )$ denotes the VAE likelihood function (decoder) and $\mathbf{z} = g(\boldsymbol{\lambda}, \hat{\boldsymbol{\epsilon}})$ denotes the functional sampler.  Note that the VAE has two sets of parameters: $\mathbf{z}$, the latent variable on which we place the prior, and the weights of the decoder, denoted as $\mathbf{W}$.  The weights must have some value during RP training and thus we place a standard normal prior on $\mathbf{W}$ and sample from this prior during optimization of $g$.  

Figure \ref{fig:03-03}b shows samples from the VAE's RP.  We see that the learned IP is drastically different than the standard Normal that is typically used: the IP is multimodal and has a much larger variance.  Yet, the difference is intuitive: placing most prior mass at opposite sides of the latent space encourages the VAE to space it's latent representations with as much distance as possible, ensuring they are as identifiable w.r.t. the model likelihood, the VAE decoder, as possible.  Interestingly, recent work by Hoffman \& Johnson (2016) suggests that VAEs can be improved by multimodal priors: "[T]he [VAE’s] individual encoding distributions $q(\mathbf{z}_{i} | \mathbf{x}_{i})$ do not have significant overlap$\ldots$then perhaps we should investigate multimodal priors that can meet $q(\mathbf{z})$ halfway" \cite{2016elbo}.  This suggests using multimodal, dispersed priors encourages flexibility and objectivity in the posterior distribution.  

We can also see analytically that the distribution in Figure \ref{fig:03-03}b allows the VAE `to follow the data' as a good RP should.  For simplicity, consider using a bivariate Gaussian as the RP approximation, and assuming it captures the same distribution as in Figure \ref{fig:03-03} (b), it's parameters would be approximately $\{ \boldsymbol{\mu} = \mathbf{0}, \boldsymbol{\Sigma} = 200 \ \mathbb{I}_{2 \times 2} \}$.  Next recall the VAE's optimization objective (the ELBO): $\mathcal{L}_{\text{VAE}} = \mathbb{E}_{q}[{-}\log p(\mathbf{x}|\mathbf{z})] + \text{KLD}[q(\boldsymbol{\mu}, \boldsymbol{\Sigma}) \mid \mid p(\mathbf{0}, 200 \ \mathbb{I}_{2 \times 2}) ]$.  The first term optimizes the model w.r.t. the data and the second acts as regularization, ensuring the variational posterior $q$ is close to the prior.  Assuming $q$'s covariance matrix is also diagonal, we can write $ \text{KLD}[q(\boldsymbol{\mu}, \boldsymbol{\Sigma}) \mid \mid p(\mathbf{0}, 200 \ \mathbb{I}_{2 \times 2}) ] = \text{KLD}[q(\boldsymbol{\mu}, \boldsymbol{\Sigma}) \mid \mid p(\mathbf{0}, \mathbb{I}_{2 \times 2})] - \frac{1}{2}\log 200$. This means that using the standard Normal up-weights the regularization (towards the prior) by about a factor of $\sqrt{200}$.

\section{CONCLUSIONS}
We have introduced two flexible, widely applicable, and derivation-free methods for approximating reference priors.  The first optimizes a new lower bound on the reference prior objective and allows for parametric or non-parametric approximations to be employed, depending on whether the user prefers to easily evaluate the prior or to have a maximally expressive approximation.  The second method uses a recently proposed particle technique to also allow for free-form approximations.  We demonstrated quantitatively and qualitatively that these methods can recover the true reference priors for univariate distributions as well as generalize to more exotic models such as Variational Autoencoders.  

Looking forward, we believe using similar techniques for constructing priors that optimize objectives other than mutual information presents a promising next step.  For example, Liu et al.\ (2014) showed that priors that maximize divergence measures other than KLD, such as Hellinger distance, between the prior and posterior have desirable properties.   Extending the proposed approximation techniques to these other families of objectives may enable new classes of Bayesian prior distributions.  

\subsubsection*{Acknowledgements}
This work was supported by the National Science Foundation under awards IIS-1320527 and
DGE-1633631 and by the National Institutes of Health under award NIH-1U01TR001801-01.
\nocite{liu2014divergence}
\bibliographystyle{plain}
\bibliography{iclr2017_conference}
\end{document}